# Scene-aware SAR Ship Detection Guided by Unsupervised Sea-land Segmentation

Han Ke
*School of Artificial Intelligence and Big Data*
*Chongqing Metropolitan College of Science and Technology*
Chongqing, China
2211031015@stu.cqcst.edu.cn

Xiao Ke
*School of Information and Communication Engineering*
*University of Electronic Science and Technology of China*
Chengdu, China
xke@std.uestc.edu.cn

Ye Yan*
*School of Artificial Intelligence and Big Data*
*Chongqing Metropolitan College of Science and Technology*
Chongqing, China
yanye831109@126.com
*Corresponding author

Rui Liu
*School of Artificial Intelligence and Big Data*
*Chongqing Metropolitan College of Science and Technology*
Chongqing, China
2211031005@stu.cqcst.edu.cn

Jinpeng Yang
*School of Artificial Intelligence and Big Data*
*Chongqing Metropolitan College of Science and Technology*
Chongqing, China
2211030904@stu.cqcst.edu.cn

Tianwen Zhang
*School of Information and Communication Engineering*
*University of Electronic Science and Technology of China*
Chengdu, China
twzhang@std.uestc.edu.cn

Xu Zhan
*School of Information and Communication Engineering*
*University of Electronic Science and Technology of China*
Chengdu, China
zhanxu@std.uestc.edu.cn

Xiaowo Xu
*School of Information and Communication Engineering*
*University of Electronic Science and Technology of China*
Chengdu, China
xuxiaowo@std.uestc.edu.cn

***Abstract*—DL-based Synthetic Aperture Radar (SAR) ship detection has tremendous advantages in numerous areas. However, it still faces some problems, such as the lack of prior knowledge, which seriously affects detection accuracy. In order to solve this problem, we propose a scene-aware SAR ship detection method based on unsupervised sea-land segmentation. This method follows a classical two-stage framework and is enhanced by two models: the unsupervised land and sea segmentation module (ULSM) and the land attention suppression module (LASM). ULSM and LASM can adaptively guide the network to reduce attention on land according to the type of scenes (i.e., inshore scene and offshore scene) and add prior knowledge (sea-land segmentation information) to the network, thereby reducing the network's attention to land directly and enhancing offshore detection performance relatively. This increases the accuracy of ship detection and enhances the interpretability of the model. Specifically, in consideration of the lack of land–sea segmentation labels in existing deep learning-based SAR ship detection datasets, ULSM uses an unsupervised approach to classify the input data scene into inshore and offshore types and performs sea-land segmentation for inshore scenes. LASM uses the sea-land segmentation information as prior knowledge to reduce the network's attention to land. We conducted our experiments using the publicly available SSDD dataset, which demonstrated the effectiveness of our network.**

## I. Introduction

Synthetic Aperture Radar (SAR) is an advanced remote sensing tool that uses microwave imaging to acquire images of the earth. It allows for all-day [1], all-weather [2] sensing. Unlike optical remote sensing, SAR does not rely on visible light, allowing it to operate at night and in harsh environments. Due to the numerous advantages of SAR mentioned above, it is well-suited for a wide range of marine missions, such as maritime rescue [3], marine meteorological observations, pollution management [4-7], and ship detection[8, 9]. Since ship detection is indispensable for many maritime tasks, it is an essential and vital task. There are many scholars who have carried out research on SAR ship detection. In the early period, a number of SAR ship detection methods have been proposed based on traditional approaches, the most classical of which is Constant False Alarm Rate [10] (CFAR). CFAR balances detection sensitivity and false alarm rate by adjusting a threshold value. However, it still faces the problems of high computational complexity in modeling and labor consumption. Nowadays, with the development of deep learning, more and more scholars are focusing on deep learning-based ship detection. They proposed DL-based methods for SAR ship detection and made many achievements. For example, Ke et al. [11] address the problem of lack of high-quality semantic information in shallow feature maps by establishing long-distance dependencies. Zhang et al. [12] proposed an FPN augmented by deformed convolution to deal with the problem of ship multi-scale in SAR ship detection. Chen et al. [13] proposed a model that adeptly combines atrous spatial pyramid pooling with shuffling attention. This model is optimized using the SIoU loss function to effectively tackle the multi-scale challenges of synthetic aperture radar (SAR) ships in complex contexts.

However, most current SAR ship detection methods [14-19] do not fully utilize the prior knowledge, especially sea-land segmentation information, which affects the improvement of ship detection accuracy. Therefore, we propose a scene-aware SAR ship detection method based on unsupervised sea-land segmentation. capable of generating prior knowledge (i.e., sea-land segmentation information) in unsupervised manners and leveraging it through network learning. The reason why we

This work was supported by the Chongqing Municipal Education Commission (Document No. YuJiaoGaoHan [2022] No. 36) and the Office of the Ministry of Education (Document No. JiaoGaoTingHan [2022] No. 14).

adopt unsupervised mechanism is that the lacking of sea-land segmentation label in existing deep learning-based SAR ship detection datasets [20-23].Our method follows the classical two-stage framework [24] and is enhanced by two models: the unsupervised land and sea segmentation module (ULSM) and the land attention suppression module (LASM). ULSM adopts an unsupervised mechanism (i.e., K-means[25]) to classify different scenes (i.e., inshore scenes and offshore scenes). For inshore scenes, it utilizes a traditional binarization mechanism (i.e., the Otsu method [26]) to generate sea-land segmentation mask maps. This establishes a foundation for the subsequent incorporation of prior knowledge into the neural network. After convolution layers, feature maps of inshore scenes and corresponding sea-land segmentation mask maps are inputted to LASM. LASM utilizes the sea-land segmentation mask maps as guidelines to adaptively reduce the attention of land regions in the feature maps directly and enhance the attention of sea regions relatively. The experiments using the publicly available SSDD [20] dataset demonstrated the effectiveness of our network.

## II. METHODOLOGY

### A. Overall Framework of our method

Figure.1 shows the architecture of our proposed method. We used the classical ResNet-50 for feature extraction. Two brand new modules were added before and after the ResNet-50, respectively (i.e., ULSM and LASM). Raw SAR images are used as inputs for ULSM which employs an unsupervised K-means clustering algorithm to classify the various scenes in the dataset into inshore and offshore scenes. Then, a sea-land segmentation mask map will be generated for inshore SAR images by the traditional Otsu method. After the feature extraction, the feature map of the inshore scene and its sea-land segmentation mask map are fed into the LASM. The LASM utilizes the sea-land segmentation mask maps to adaptively reduce the attention of land region.

### B. Unsupervised land and sea segmentation module

Due to the lack of sea–land segmentation labels in existing deep learning-based SAR ship detection datasets, we propose an unsupervised land and sea segmentation module (ULSM) to perform sea–land segmentation without supervision. In ULSM, two algorithms are utilized: k-means clustering and Otsu method The k-means clustering algorithm is employed to distinguish between inshore and offshore scenes within the dataset. We began by randomly initializing two centroids for the clustering process. Subsequently, ResNet50 was used to extract features from the data, resulting in feature vectors. For each feature vector $v_i$, K-means calculates its distance from the two cluster centers and assigns each feature vector to the category of the nearest centroid. The centroid of each cluster is then recalculated by averaging all feature vectors assigned to that cluster. This process is repeated until the centroids stabilize or the maximum number of iterations is reached. The distance calculating process is as follows:

$$d_{i1} = \sqrt{(v_{i1} - c_{11})^2 + (v_{i2} - c_{12})^2 + \cdots + (v_{in} - c_{1n})^2} \quad (1)$$

where $d_{i1}$ denotes the distance of feature vector $v_i$ from centroid 1, $v_{in}$ denotes the value of the nth dimension of the feature vector, and $c_{1n}$ denotes the value of the nth dimension of the centroid.

After obtaining the clustering results, we consider the smaller cluster as the inshore scene and the larger one as the offshore scene. This is based on two reasons: Firstly, the number of samples for inshore scenes is much smaller than that for offshore scenes in most existing SAR ship datasets. Secondly, in reality, the ocean area is larger than the land area on Earth's surface [27].

After completing k-means clustering, we will use the Otsu method to segment the images of inshore scenes into land and sea regions. The Otsu method determines an optimal threshold by maximizing inter-class variance, which segments the image into foreground and background regions (i.e., sea and land

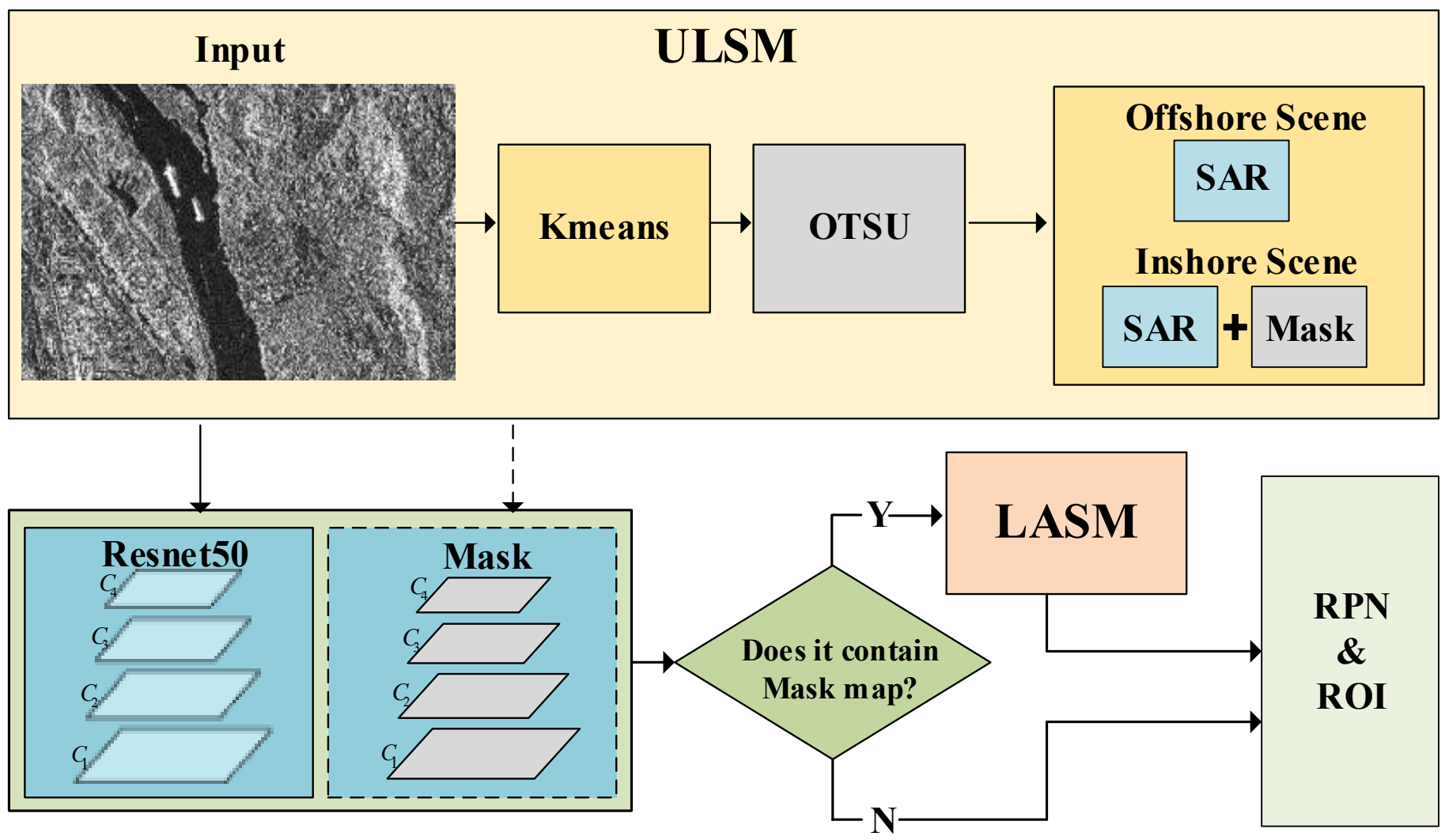

Fig. 1. Overall Framework of the proposed scene-aware method based on unsupervised sea-land segmentation

regions). The process of calculating the grayscale mean of all pixels in an image is as follows:

$$\mu T = \sum_{i=0}^{255} i \cdot p_i \tag{2}$$

where $\mu T$ denotes the average grayscale value of all pixels in the image, $i$ denotes the gray level of the image, $p_i$ denotes the probability of gray level $i$ appearing in the image, and $i \cdot p_i$ denotes the contribution of gray level $i$ to the global mean. The process of selecting the optimal threshold consists of two steps. We will first calculate the between-class variance for each possible threshold, as shown:

$$\sigma^2(k) = \frac{(\mu T \cdot \omega(k) - \mu(k))^2}{\omega(k) \cdot (1 - \omega(k))} \tag{3}$$

The inter-class variance ($\sigma^2(k)$) measures the degree of separation between the two classes (sea and land regions) when the image is segmented using a threshold k. Here, k represents the current candidate threshold, and $\omega(k)$ denotes the cumulative probability of a pixel being below threshold k. Finally, the optimal threshold is determined by selecting the k that maximizes the inter-class variance ($\sigma^2(k)$). The process is as follows:

$$k^* = \arg\max_k \sigma^2(k) \tag{4}$$

We set pixels with grayscale values below the threshold to 0 (indicating land regions) and pixels with grayscale values above the threshold to 1 (indicating sea regions).

Moreover, it is worth noticed that we choose unsupervised methods to classify the type of scene instead of supervised methods. The reason is that the supervised methods necessitate labeled datasets, which are scarce, and acquiring coastline data can be particularly challenging. Therefore, we employed two unsupervised techniques, k-means clustering and the Otsu method, to segment the inshore scene SAR images between land and sea regions.

Furthermore, we opted to combine the unsupervised methods of K-means and the Otsu method for sea-land segmentation instead of relying solely on the Otsu method. This is because the Otsu method's segmentation results tend to be coarse, and the inshore scenes do not require sea-land segmentation. Using the Otsu method alone for the entire dataset could lead to misclassifying large ships as land, resulting in inaccurate sea-land segmentation masks that hinder the network's correct utilization of prior information. By incorporating K-means to classify both inshore and offshore scenes, this issue can be mitigated. The generated sea-land segmentation mask map will be used to guide the land attention suppression module to adaptively reduce the attention of land regions in the feature maps.

### *C. Land attention suppression module.*

Figure 2 illustrates the architecture of the Land Attention Suppression Module (LASM). The LASM utilizes the feature map and its corresponding sea-land segmentation mask map to adaptively suppress the neural network's attention to land regions. The outputs of LASM are sent to the RPN for candidate frame generation and subsequently to the ROI for target classification and precise localization. The feature maps of varying dimensions are processed through a 1×1 convolution to standardize them to a uniform dimension of 256. Next, the feature maps are processed through average pooling to form 7×7×256 feature vectors, which are then flattened into 1-dimensional vectors of size 12544. These vectors pass through the fully connected layer twice, followed by the Tanh activation function applied twice. The resulting output is then divided by 2 to yield the final land attention suppression weights. The process of weight map generating is as follows:

$$A = M \cdot (1 - \lambda) + (\mathrm{I} - M) \cdot 1 \tag{5}$$

where $A$ denotes the attention map, I denotes a matrix in which every element has a value of 1, and it shares the same dimensions as $M$. $M$ signifies the sea-land segmentation mask map, and $\lambda$ represents the land attention suppression weights. The generated weight maps are upsampled to match the dimensions of the corresponding hierarchical feature maps. These weight maps are then matrix-multiplied with the initial feature maps to reduce the attention to land regions within the initial feature maps. The reason we apply the sea-land segmentation mask map at the feature level, rather than filtering shore targets directly at the detection result level, is due to the limited accuracy of unsupervised methods. Applying the mask map directly at the detection result level may inadvertently filter out ground truth boxes, thus reducing detection accuracy.

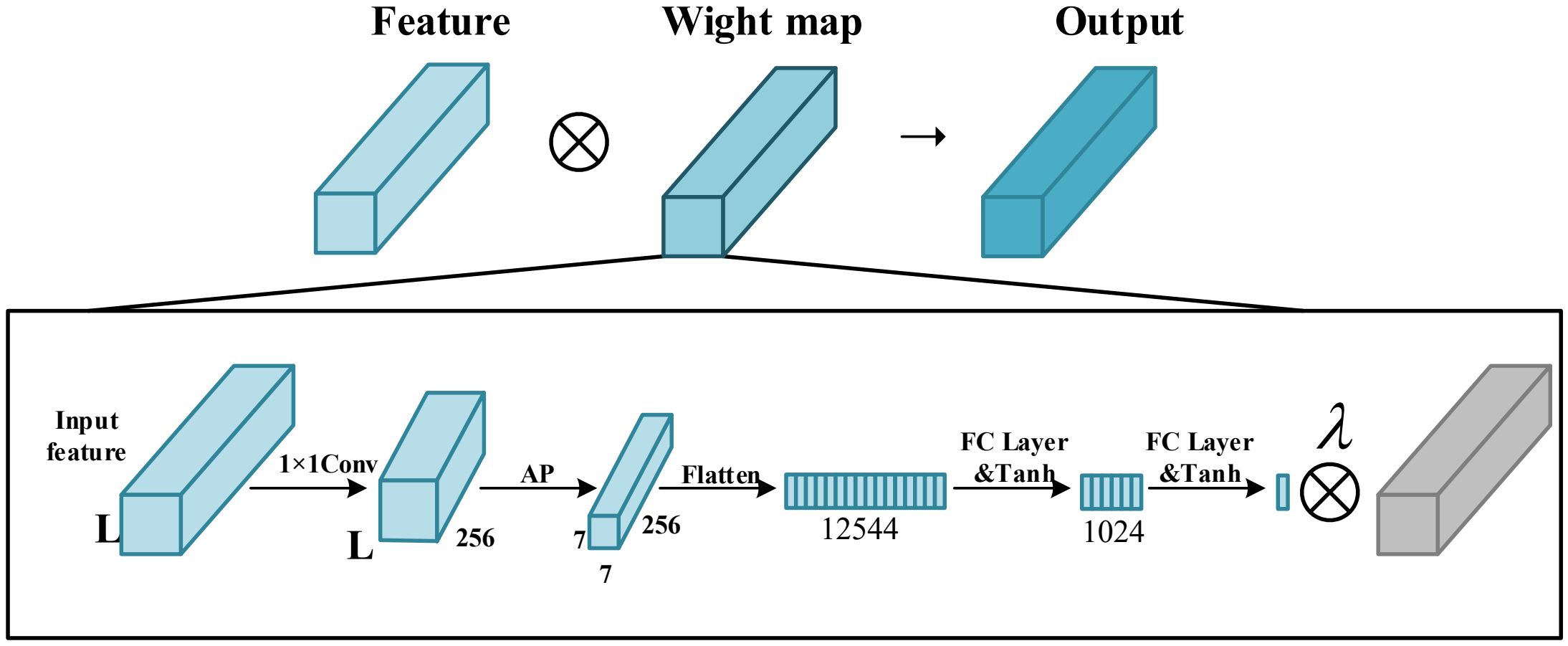

Fig. 2. The architecture of land attention suppression module

## III. EXPERIMENTS

### A. *Dataset*

The SSDD [20] dataset, specifically designed for ship detection in remotely sensed imagery, encompasses a wide range of sea states, lighting conditions, ship types, and sizes. Each image is meticulously labeled by hand to ensure the accuracy of the ship's position and size. As an open-access resource, the SSDD dataset allows researchers to easily download and utilize it, thereby promoting scholarly communication and advancement. Due to these advantages, many scholars have used the SSDD dataset in their experiments. In this paper, we also utilize the SSDD dataset for our experiments. The results of these experiments demonstrate the effectiveness of our proposed method.

### B. *Training Details*

In this paper, we crop the images to a size of 512×512 pixels. We employ the classic two-stage algorithm framework for our approach, utilizing ResNet-50 for feature extraction. The learning rate is set to 0.02, momentum to 0.9, and weight decay to 0.001. We train our network for 12 epochs using a personal computer equipped with an RTX 3060 GPU and an AMD Ryzen 6000 series CPU.

## IV. RESULT

### A. *Qualitative and Quantitative Experimental Results*

Table I presents the quantitative evaluation results of our proposed method and three other methods for detection. We compared our scene-aware method based on unsupervised sea-land segmentation. with Faster-RCNN [24], DCN [28], and Grid-RCNN [29]. As Table I shows, our method achieves higher mAP and mAP50. Specifically, the mAP50 of our method is 91.8%, which is approximately 2.6% higher than the second highest model, DCN. Furthermore, our method also achieves the highest mAPS, indicating its excellent performance on small ships. These results demonstrate the effectiveness of our method in SAR ship detection.

TABLE I. DETECTION RESULTS

| METHOD | MAP | MAP$_{50}$ | MAP$_{75}$ | MAP$_{S}$ | MAP$_{M}$ | MAP$_{L}$ |
|---|---|---|---|---|---|---|
| OURS | 63.5% | 91.8% | 75.2% | 63.8% | 65.1% | 23.1% |
| FASTER-RCNN | 62.5% | 89.0% | 75.4% | 61.9% | 66.1% | 39.0% |
| DCN | 63.0% | 89.8% | 75.8% | 63.1% | 65.2% | 38.0% |
| GRID-RCNN | 62.5% | 86.5% | 75.1% | 60.8% | 68.7% | 40.7% |

Figure 3 showcases the detection results of our proposed method alongside those of the second-best method, DCN. In this figure, the first row of images (i.e., samples a, b, and c) presents the visualization results of our approach, whereas the second row (i.e., samples d, e, and f) displays the outcomes of the second-best method, DCN, on the same images. The white boxes represent the ground truth, while the yellow boxes indicate the bounding boxes. As illustrated in Figure 3, our method not only achieves higher detection accuracy but also significantly reduces the occurrence of false negatives.

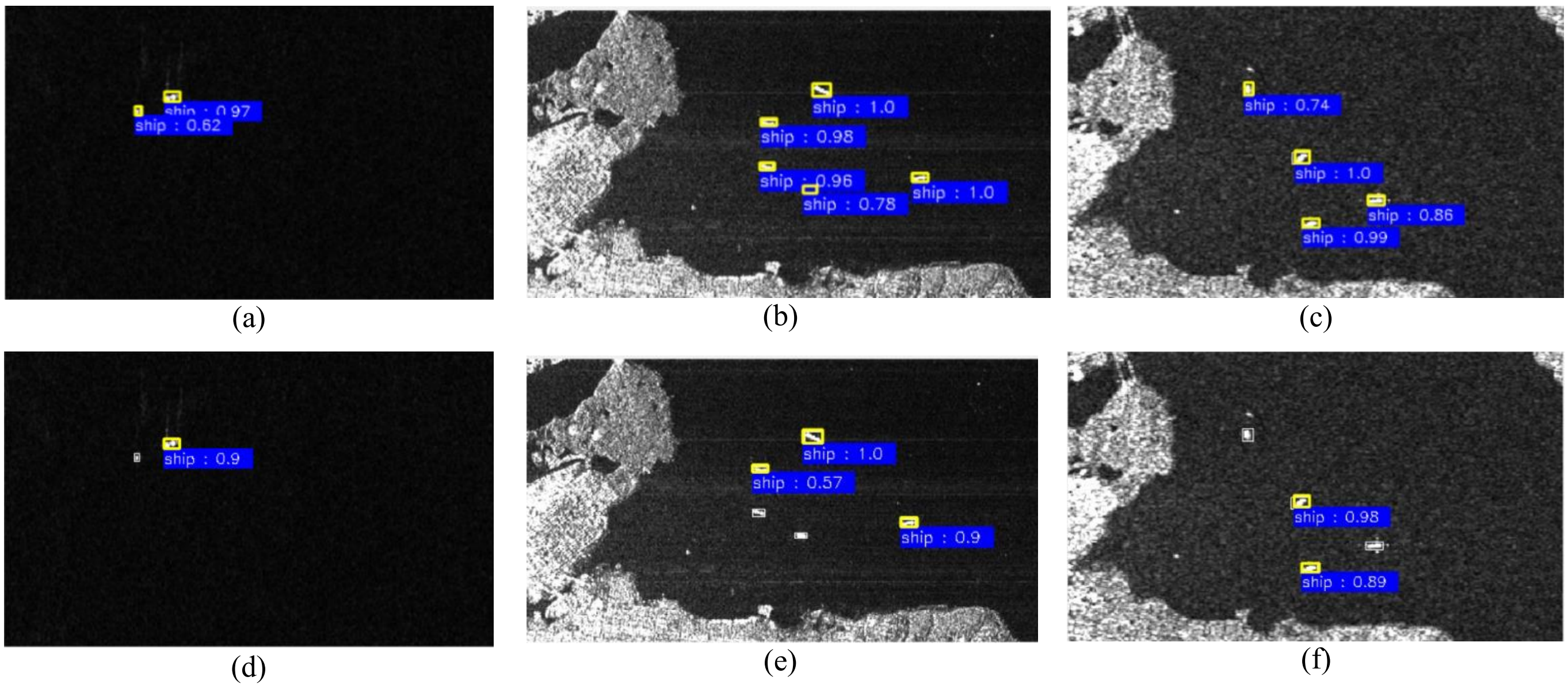

Fig. 3. The detection results of our method and DCN

Meanwhile, as illustrated in Figure 3, our method performs well in both inshore scenes (e.g., sample b and sample c) and offshore scenes (e.g., sample a). For instance, as shown in sample b, our method is effective even in complex scenes. According to the above quantitative evaluation results, the scene-aware method based on unsupervised sea-land segmentation achieves good SAR ship detection performance.

### *B. Ablation Study*

Table II demonstrates the impact of utilizing the K-means clustering algorithm to classify the dataset into inshore and offshore scenes prior to applying the Otsu method for sea-land segmentation in the ULSM. ID-1 represents our complete method, whereas ID-2 represents the application of the Otsu method alone on the entire dataset for sea-land segmentation without the inclusion of K-means clustering. The results clearly indicate that incorporating the K-means clustering algorithm in our method significantly enhances both mAP and $mAP_{50}$ scores. This demonstrates the effectiveness of performing the K-means operation prior to applying the Otsu method. The decrease in mAP and $mAP_{50}$ in ID-2 is due to the offshore scenes don't require sea-land segmentation. This leads to large ships being misclassified as land and creating inaccurate masks, which hinders the network's utilization of prior information.

TABLE II. ABLATION STUDY ON OUR METHOD

| ID | K-MEANS | MAP | $MAP_{50}$ |
|---|---|---|---|
| **1** | ✓ | **63.5%** | **91.8%** |
| **2** | X | 62.2% | 88.4% |

## V. CONCLUSIONS

In this paper, we propose a scene-aware method based on unsupervised sea-land segmentation, enhanced by two modules: the Land Attention Suppression Module (LASM) and the Unsupervised Land and Sea Segmentation Module (ULSM). ULSM generates prior knowledge (i.e., sea-land segmentation information) according to different scenes (i.e., inshore and offshore) in unsupervised manners. LASM leverages this information to directly reduce the network's attention to land in the feature level, and relatively enhancing offshore detection performance. Ablation studies confirm the effectiveness of performing the K-means operation prior to applying the Otsu method. And the conducted comparative experiments on SSDD show that our method achieves good SAR ship detection performance.

## ACKNOWLEDGMENT

This work was supported by the Chongqing Municipal Education Commission (Document No. YuJiaoGaoHan [2022] No. 36) and the Office of the Ministry of Education (Document No. JiaoGaoTingHan [2022] No. 14). The funding and resources provided were crucial to the successful completion of this study, and we express our gratitude for their support.

## REFERENCES

[1] Yunkai, D., et al., *Forthcoming spaceborne SAR development.* JOURNAL OF RADARS, 2020. **9**(1): p. 1-33.

[2] Yun-kai, D., Z. Feng-jun, and W. Yu, *Brief analysis on the development and application of spaceborne SAR.* JOURNAL OF RADARS, 2012. **1**(1): p. 1-10.

[3] Asiyabi, R.M., et al., *Synthetic aperture radar (SAR) for ocean: A review.* IEEE Journal of Selected Topics in Applied Earth Observations and Remote Sensing, 2023.

[4] Migliaccio, M., F. Nunziata, and A. Buono, SAR polarimetry for sea oil slick observation. International Journal of Remote Sensing, 2015. 36(12): p. 3243-3273.

[5] Mohammadi, M., et al., *Detection of oil pollution using SAR and optical remote sensing imagery: A case study of the Persian Gulf.* Journal of the Indian Society of Remote Sensing, 2021. **49**(10): p. 2377-2385.

[6] Gade, M. and W. Alpers, *Using ERS-2 SAR images for routine observation of marine pollution in European coastal waters.* Science of the total environment, 1999. **237**: p. 441-448.

[7] Trivero, P. and W. Biamino, *Observing marine pollution with synthetic aperture radar.* Geoscience and remote sensing new achievements, 2010: p. 397-418.

[8] Li, J., et al., *Deep learning for SAR ship detection: Past, present and future.* Remote Sensing, 2022. **14**(11): p. 2712.

[9] Yasir, M., et al., *Ship detection based on deep learning using SAR imagery: a systematic literature review.* Soft Computing, 2023. **27**(1): p. 63-84.

[10] Gandhi, P.P. and S.A. Kassam, *Analysis of CFAR processors in nonhomogeneous background.* IEEE Transactions on Aerospace and Electronic systems, 1988. **24**(4): p. 427-445.

[11] Ke, X., et al. *Sar ship detection based on swin transformer and feature enhancement feature pyramid network*. in *IGARSS 2022-2022 IEEE International Geoscience and Remote Sensing Symposium*. 2022. IEEE.

[12] Zhang, Z.T., X. Zhang, and Z. Shao. *Deform-FPN: A Novel FPN with Deformable Convolution for Multi-Scale SAR Ship Detection*. in *IGARSS 2023-2023 IEEE International Geoscience and Remote Sensing Symposium*. 2023. IEEE.

[13] Chen, Z., et al., *Multi-scale ship detection algorithm based on YOLOv7 for complex scene SAR images.* Remote Sensing, 2023. **15**(8): p. 2071.

[14] Zhang, T., X. Zhang, and X. Ke, *Quad-FPN: A novel quad feature pyramid network for SAR ship detection.* Remote Sensing, 2021. **13**(14): p. 2771.

[15] Chang, Y.-L., et al., *Ship detection based on YOLOv2 for SAR imagery.* Remote Sensing, 2019. **11**(7): p. 786.

[16] Zhang, T., et al., *HOG-ShipCLSNet: A novel deep learning network with hog feature fusion for SAR ship classification.* IEEE Transactions on Geoscience and Remote Sensing, 2021. **60**: p. 1-22.

[17] Ke, X., et al. *SAR ship detection based on an improved faster R-CNN using deformable convolution*. in *2021 IEEE International Geoscience and Remote Sensing Symposium IGARSS*. 2021. IEEE.

[18] Ke, H., et al. *Laplace & LBP feature guided SAR ship detection method with adaptive feature enhancement block*. in *2024 IEEE 6th Advanced Information Management, Communicates, Electronic and Automation Control Conference (IMCEC)*. 2024. IEEE.

[19] Zhang, T. and X. Zhang, *High-speed ship detection in SAR images based on a grid convolutional neural network.* Remote Sensing, 2019. **11**(10): p. 1206.

[20] Zhang, T., et al., *SAR ship detection dataset (SSDD): Official release and comprehensive data analysis.* Remote Sensing, 2021. **13**(18): p. 3690.

[21] Zhang, T., et al., *LS-SSDD-v1. 0: A deep learning dataset dedicated to small ship detection from large-scale Sentinel-1 SAR images.* Remote Sensing, 2020. **12**(18): p. 2997.

[22] Wei, S., et al., *HRSID: A high-resolution SAR images dataset for ship detection and instance segmentation.* Ieee Access, 2020. **8**: p. 120234-120254.

[23] Xian, S., et al., *AIR-SARShip-1.0: High-resolution SAR ship detection dataset.* JOURNAL OF RADARS, 2019. **8**(6): p. 852-863.

[24] Ren, S., et al., *Faster r-cnn: Towards real-time object detection with region proposal networks.* Advances in neural information processing systems, 2015. **28**.

[25] Krishna, K. and M.N. Murty, *Genetic K-means algorithm.* IEEE Transactions on Systems, Man, and Cybernetics, Part B (Cybernetics), 1999. **29**(3): p. 433-439.

[26] Zhang, J. and J. Hu. *Image segmentation based on 2D Otsu method with histogram analysis*. in *2008 international conference on computer science and software engineering*. 2008. IEEE.

[27] Zhang, T., et al., *Balance scene learning mechanism for offshore and inshore ship detection in SAR images.* IEEE geoscience and remote sensing letters, 2020. **19**: p. 1-5.

[28] Dai, J., et al. *Deformable convolutional networks*. in *Proceedings of the IEEE international conference on computer vision*. 2017.

[29] Lu, X., et al. *Grid r-cnn*. in *Proceedings of the IEEE/CVF Conference on Computer Vision and Pattern Recognition*. 2019.